\documentclass[final,5p,times,twocolumn,authoryear]{elsarticle}

\usepackage{amssymb}
\usepackage{amsmath}
\usepackage{array}
\usepackage{algorithmic}
\usepackage{graphicx}
\usepackage{booktabs}
\usepackage{multirow}
\usepackage{pifont}
\usepackage{makecell}
\usepackage[table,xcdraw]{xcolor} 
\usepackage{algorithm}
\usepackage{color}
\usepackage{xcolor}
\usepackage{hyperref}
\journal{Medical Image Analysis}

\begin{document}

\begin{frontmatter}



\title{Rethinking Data Imbalance in Class Incremental Surgical Instrument Segmentation} 

\author[CUHK]{Shifang Zhao\fnref{eqcon}}
\ead{shifang.zhao@bjtu.edu.cn}

\author[CUHK,TUM]{Long Bai\fnref{eqcon}\corref{cor1}}
\ead{b.long@link.cuhk.edu.hk}

\author[TUM,US,IHU]{Kun Yuan}
\ead{kun.yuan@tum.de}

\author[TUM]{Feng Li}
\ead{feng.li@tum.de}

\author[CUHK]{Jieming Yu}
\ead{jmyu@link.cuhk.edu.hk}

\author[CUHK]{Wenzhen Dong}
\ead{dongwz@link.cuhk.edu.hk}

\author[CUHK]{Guankun Wang}
\ead{gkwang@link.cuhk.edu.hk}

\author[UCL]{Mobarakol Islam}
\ead{mobarakol.islam@ucl.ac.uk}

\author[US,IHU]{Nicolas Padoy}
\ead{npadoy@unistra.fr}

\author[TUM]{Nassir Navab}
\ead{nassir.navab@tum.de}

\author[CUHK]{Hongliang Ren\corref{cor1}}
\ead{hlren@ee.cuhk.edu.hk}

\fntext[eqcon]{Co-first authors.}
\cortext[cor1]{Corresponding author.}
\affiliation[CUHK]{organization={Department of Electronic Engineering, The Chinese University of Hong Kong},
            city={Hong Kong},
            country={Hong Kong}}

\affiliation[TUM]{organization={Chair for Computer Aided Medical Procedures, Technical University of Munich},
            city={Munich},
            country={Germany}}

\affiliation[US]{organization={University of Strasbourg, CNRS, INSERM, ICube},
            city={Strasbourg},
            country={France}}

\affiliation[IHU]{organization={IHU Strasbourg},
            city={Strasbourg},
            country={France}}
            
\affiliation[UCL]{organization={UCL Hawkes Institute, University College London},
            city={London},
            country={United Kingdom}}

\begin{abstract}
In surgical instrument segmentation, the increasing variety of instruments over time poses a significant challenge for existing neural networks, as they are unable to effectively learn such incremental tasks and suffer from catastrophic forgetting. When learning new data, the model experiences a sharp performance drop on previously learned data. Although several continual learning methods have been proposed for incremental understanding tasks in surgical scenarios, the issue of data imbalance often leads to a strong bias in the segmentation head, resulting in poor performance. Data imbalance can occur in two forms: (i) class imbalance between new and old data, and (ii) class imbalance within the same time point of data. Such imbalances often cause the dominant classes to take over the training process of continual semantic segmentation (CSS). To address this issue, we propose \textbf{SurgCSS}, a novel plug-and-play CSS framework for surgical instrument segmentation under data imbalance. Specifically, we generate realistic surgical backgrounds through inpainting and blend instrument foregrounds with the generated backgrounds in a class-aware manner to balance the data distribution in various scenarios. We further propose the Class Desensitization Loss by employing contrastive learning to correct edge biases caused by data imbalance. Moreover, we dynamically fuse the weight parameters of the old and new models to achieve a better trade-off between the biased and unbiased model weights. To investigate the data imbalance problem in surgical scenarios, we construct a new benchmark for surgical instrument CSS by integrating four public datasets: EndoVis 2017, EndoVis 2018, CholecSeg8k, and SAR-RAPR50. Extensive experiments demonstrate the effectiveness of the proposed framework, achieving significant performance improvement against existing baselines. Our method demonstrates excellent potential for clinical applications. The code is publicly available at \href{https://github.com/Zzsf11/SurgCSS}{github.com/Zzsf11/SurgCSS}.
\end{abstract}


\end{frontmatter}

\section{Introduction}
\label{sec:introduction}
\begin{figure*}[!t]
\centering
\includegraphics[width=1\linewidth]{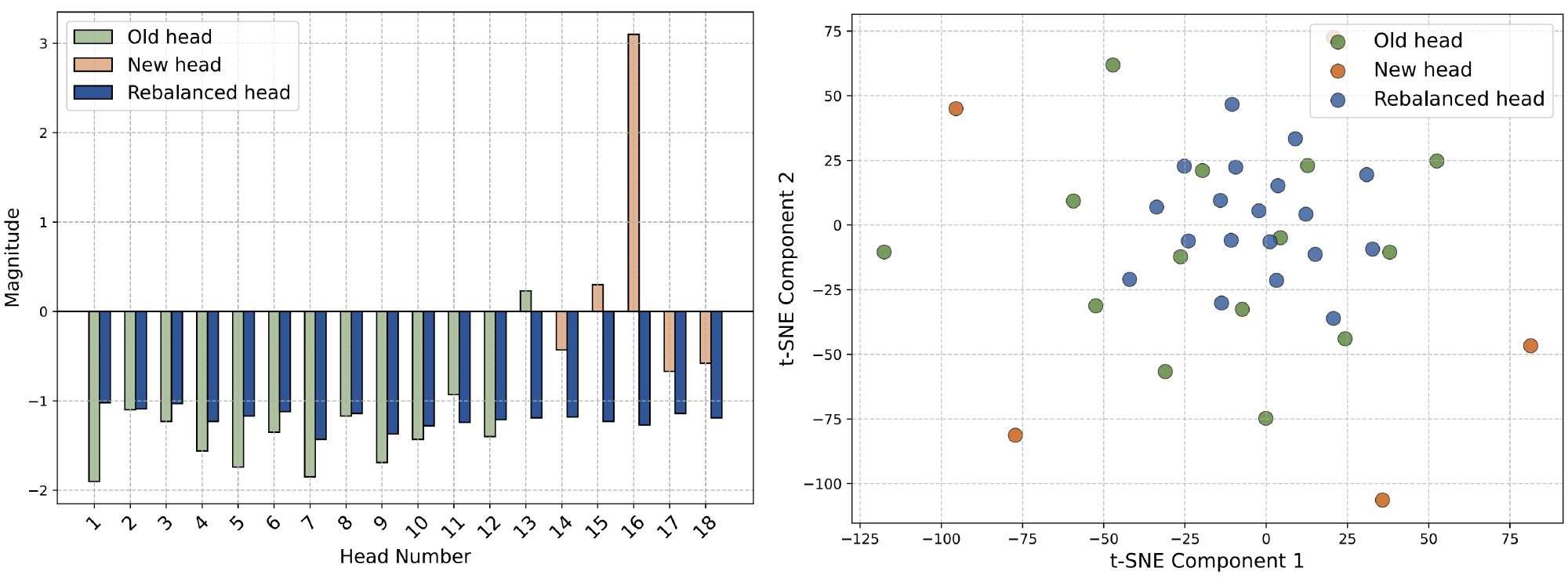}
\caption{Experimental results on a surgical dataset confirm the presence of bias in the final layer, as observed through weight visualization of the segmentation head in task ``13-1". Heads 1-13 correspond to the old class, while heads 14-18 represent incremental classes, one by one (Best viewed in color). We visualize the bias within each class-specific output channel of the final convolutional layer using two methods: a summation average of activations (left) and a t-SNE embedding of the feature space (right).
}
\label{fig:biased_head}
\end{figure*}

Image segmentation of surgical instruments is crucial for advancing fully autonomous robotic surgery, as it provides critical information to understand dynamic and complex surgical scenes~\citep{chen2024asi,nwoye2023cholectriplet2022,maier2022surgical,cheng2025mamba,yu2024sam, ranem2024continual}. Recently, deep learning-based methods have shown promising results in solving increasingly challenging tasks~\citep{garcia2021image,colleoni2022ssis,lou2023min,liu2021graph,alabi2025multitask,yu2024adapting}. However, collecting and aggregating a large-scale dataset for surgical scenarios is difficult in practice due to high storage costs, licensing, and privacy concerns, and the frequent updating of instruments~\citep{zia2023surgical,razzak2018deep,wang2023rethinking,bai2023revisiting}. Most approaches suffer from \textit{catastrophic forgetting}, where the performance has a significant degradation when the past data are not available. To address this challenge, Continual Learning (CL) methods aim to enable models to learn new classes or tasks by training on new data while alleviating the catastrophic forgetting problem~\citep{chaudhry2018riemannian, zhao2024transferring,li2017learning,douillard2020podnet,li2025clms,ayromlou2024ccsi,wang2023domain}. Nevertheless, previous methods suffer significant performance degradation in robotic surgical scenarios due to biased incremental learning updates, as illustrated in Figure~\ref{fig:biased_head}, which shows a pronounced bias in the segmentation head weights for new and old classes.

\textit{Why is continual segmentation more challenging in robotic surgical scenarios?} We believe this stems from the combination of three factors: (i) the data distribution for instrument classes is highly imbalanced due to variations in surgical workflow, with different phase durations and erratic occurrences; (ii) the imbalance is exacerbated as the new class has a high probability of occurrence and is highly concentrated, whereas the old class has a low probability of occurrence and is more dispersed; and (iii) the visual similarity of instruments, with minimal differences near class boundaries, further complicates segmentation. Even worse, the unavailability of exemplar samples due to licensing and privacy constraints makes the task even more challenging. Additionally, attributes specific to surgical scenes lead to increased difficulty. Unlike the relatively stable illumination in general settings, surgical illumination is highly directional and often positioned to highlight specific areas. The anatomical complexity also leads to various illumination degradation scenes. The diversity of surgical instruments and the similar visual characteristics also introduce further complexity. Besides, in surgical scenes, camera viewpoints are often highly constrained, leading to noise and reduced image quality, which further increases the difficulty of applying segmentation models.

To address the class imbalance challenges, classical techniques have been widely explored in machine learning to mitigate imbalanced distributions. Recent studies~\citep{he2021tale} have analyzed the relationship between class incremental learning and class imbalanced learning, demonstrating that techniques from class imbalanced learning are similar to class incremental learning methods. \citep{wu2019large, liu2022long, zhao2020maintaining} have attributed the performance degradation caused by sample imbalance to biases in the final fully connected layer. These biases are mitigated by incorporating a bias correction layer trained using a balanced validation set ~\citep{wu2019large} or normalizing the output weights~\citep{zhao2020maintaining}. While these methods provide some improvement, they fall short in the context of continual learning for surgical instrument segmentation, where data imbalance is further exacerbated by privacy restrictions and the inherent complexity of surgical workflows. Current continual semantic segmentation approaches still lack effective strategies to improve performance under imbalanced distribution scenarios, which are typical in surgical environments. 

In medical image analysis, data scarcity is often addressed by generating synthetic data, enabling knowledge sharing without exposing patient-level information~\citep{colleoni2021robotic, colleoni2022ssis}. Similarly, continual learning utilizes generative models to replay synthesized data~\citep{shin2017continual, wu2018memory}. However, updating generative models incrementally introduces challenges, including high resource consumption and reduced system stability. In this case, Xu~\citep{xu2024privacy} blended and augmented the foreground to the synthetic background to avoid generating model updates. These methods provide the flexibility to create a class-balanced dataset using any preferred instrument, while also automatically generating the instrument annotations. However, they simply crop patches without instruments from the images and use these patches to train a generative model. In this case, (i) the generated backgrounds are often distorted and may contain unforeseen artifacts and inconsistencies, (ii) the model can only learn foreground information based on segmentation annotations from such data, but due to the distorted backgrounds, it is challenging for the model to extract effective discriminative information between the foreground and the background, and (iii) this approach limits the model to generated background patches, neglecting the effective utilization of the scarce real samples available. Therefore, although this method is straightforward, there is still significant room for improvement.

In this work, we propose \textbf{SurgCSS}, a novel plug-and-play approach specifically designed to mitigate the degradation caused by class imbalance in the continual segmentation of robotic surgical scenes. We identify a significant bias in the segmentation head, analyze its root causes, and propose a two-stage pipeline to resolve this issue. Specifically, we decouple the instrument foreground from the surgical background and construct a small class-balanced set using a class-aware blending technique, which is crucial in overcoming data scarcity in surgical scenarios. Additionally, we design the Class Desensitization Loss (CDL) function to reduce sensitivity to class boundary uncertainty, thereby minimizing the impact of visual similarity between classes. After training the model on the small set, we apply a dynamic weight fusion (DWF) strategy to merge the representation capabilities of the trained and untrained models effectively.
Our approach effectively addresses the class imbalance between new and old data, as well as the class imbalance within the same time point.
We summarize our contributions as follows: 
\begin{itemize} 
\item We propose a novel plug-and-play framework to address performance degradation in class incremental surgical instrument segmentation, specifically targeting imbalanced distributions under data-scarce surgical scenarios. 
\item We introduce a class-aware blending with an inpainting mechanism to decouple the instrument foreground from the high-fidelity surgical background, effectively tackling the data scarcity issue. The CDL is further proposed to minimize confusion between visually similar classes and correct segmentation boundary deviations. 
\item We develop a DWF strategy that combines the strengths of both incremental and rebalance models, ensuring detailed class representations are preserved while boosting overall segmentation accuracy. 
\item By integrating four existing public datasets, we establish a new large-scale benchmark for class incremental segmentation of surgical instrument segmentation. Through extensive comparisons and ablation studies on this benchmark, we demonstrate the potential of our method for clinical applications.

\end{itemize}

\section{Related Work}
\subsection{Class-Incremental Continual Learning}
Class-Incremental Continual Learning aims for a model to learn distinguishing features for new classes while preventing catastrophic forgetting of previously learned classes. This requires balancing the need for learning flexibility with maintaining memory stability~\citep{wang2024comprehensive}. In Continual Semantic Segmentation, a specific challenge is background shift, which happens when pixels labeled as background correspond to classes the model has already seen or will see later. Early methods like MiB~\citep{cermelli2020modeling} addressed this by calibrating the loss for background pixels using prior model predictions. PLOP~\citep{douillard2021plop} used a multi-scale pooling approach for distillation. SSUL~\citep{cha2021ssul} used auxiliary data to improve distillation and prevent forgetting. Other approaches focused on preventing feature changes; RCIL~\citep{zhang2022representation} and EWF~\citep{xiao2023endpoints} worked to reduce feature plasticity problems by combining parameters from the new and old models. CIT~\citep{ge2024cit} converts the output of existing semantic segmentation models into a form that does not depend on the specific class. IPSeg~\citep{yu2025ipseg} uses image posterior probabilities to align the optimization process across different stages and reduce negative effects from optimizing each stage separately. MoDA~\citep{yang2024continual} helps the SAM encoder create features that are clearly separated for different task types, providing accurate task-specific information for continual learning. EIR~\citep{yin2025beyond} reduces background changes in new images by combining stored examples with the new images. CoMBO~\citep{fang2025combo} adds an extra part to handle learning new classes while preserving the original information used for distillation. These recent methods represent the current advances in general continual segmentation scenarios.

Continual segmentation is also being explored in medical image analysis. \citep{xu2024privacy} uses synthetic data with class-aware temperature normalization to enhance model rigidity while preserving privacy in robotic surgery. ~\citep{zhang2023continual} propose replacing the conventional output layer with a set of lightweight, class-specific heads. ~\citep{ji2023continual} demonstrates that continually training, followed by freezing the encoder, combined with incrementally added decoders, can effectively extract sufficiently representative image features for new classes, allowing them to be validly segmented. ~\citep{elskhawy2020continual} introduces an approach that disentangles the feature space into task-specific and task-invariant features. \citep{zhu2024low} employs lightweight low-rank adaptation (LoRA) to efficiently extend pre-trained segmentation models for segmenting new organs. \citep{sadegheih2025modality} introduces a mixture-of-experts mechanism and dual knowledge distillation for brain lesion segmentation. \citep{anand2024label} proposes the construction of a common label space to facilitate incremental learning across different datasets. ~\citep{guo2025continual} introduces segmentation models that can handle a comprehensive set of fine-grained whole-body anatomies from diverse, partially labeled datasets in CT segmentation.

However, previous continual segmentation methods that use distillation may struggle with visual similarities between different instruments or anatomical structures, particularly near class boundaries. Methods focused on parameter fusion or regularization may not inherently address the severe class imbalance. Additionally, while data limitations represent a significant challenge in medical continual learning tasks, previous research in medical image analysis has primarily focused on modifying model architectures to improve performance, often neglecting the critical issue of data scarcity and imbalance. In this paper, we further explore the specific challenges and solutions related to surgical instrument semantic segmentation, a domain marked by imbalanced data, privacy concerns, and the complexity of surgical workflows.

\subsection{Learning with Class Imbalance}
Class imbalance in deep learning is a significant challenge arising from the disproportionate representation of classes in training datasets, which often leads to biased models favoring the majority class. To address this issue, various techniques have been proposed, broadly categorized into data-level, algorithm-level, and hybrid methods. \textit{Data-level methods}~\citep{hensman2015impact,lee2016plankton} focus on modifying the dataset to reduce class imbalance by adjusting the training data distribution, either through duplicating samples from the minority class or removing samples from the majority class. In contrast, \textit{algorithm-level approaches}~\citep{lin2017focal,wang2016training} involve modifying the learning process or the loss function to account for class imbalance.
Recently, researchers have increasingly focused on addressing class imbalance in continual learning. For instance, \citep{he2021tale} examined the relationship between class imbalance and performance degradation in continual learning. \citep{wu2019large} proposed a bias correction method that targets the fully connected layer, addressing the bias caused by class imbalance and effectively narrowing the performance gap between old and new classes. Furthermore, \citep{liu2022long} tackled class imbalance in long-tailed datasets by introducing a learnable weight-scaling layer to mitigate bias. This approach is particularly relevant to continual learning scenarios and aligns closely with our task of semantic segmentation for surgical instruments.

\subsection{Image Synthesis}
Surgical image scarcity, driven by privacy concerns and workflow complexities, challenges deep learning application development. To address this, synthetic datasets blend open-source surgical instrument images with real or simulated backgrounds, enabling targeted data generation~\citep{wang2022rethinking, garcia2021image}. However, these methods face limitations such as manual data collection costs and limited source availability. Generative models, particularly GANs, have improved data augmentation by synthesizing surgical images~\citep{zaffino2021open, shin2018medical, hamghalam2020high, lee2019davincigan, rivoir2021long}, and unpaired image-to-image translation has enhanced image quality~\citep{pfeiffer2019generating}. Recently, diffusion models (DMs) have demonstrated superior performance, employing semantic maps for high-quality surgical image synthesis~\citep{zhou2024image} and latent consistency-distilled approaches for unpaired translation~\citep{venkatesh2024surgical}. Advanced methods now generate high-resolution surgical videos from text prompts alone~\citep{cho2024surgen}. Compare to other data augmentation methods like copypaste~\citep{ghiasi2021simple} and cutpaste~\citep{li2021cutpaste}, image synthesis methods offer the capability to generate entirely new visual contexts and object appearances, create variations that extend beyond basic geometric or color transformations, and potentially produce more realistic and diverse data that better reflects the complexity of surgical environments. Our inpainting-based approach further addresses data limitations by generating realistic, class-balanced datasets.

\section{Methodology}

\begin{figure*}[!t]
\centering
\hspace{-0.3cm} 
\includegraphics[width=\linewidth]{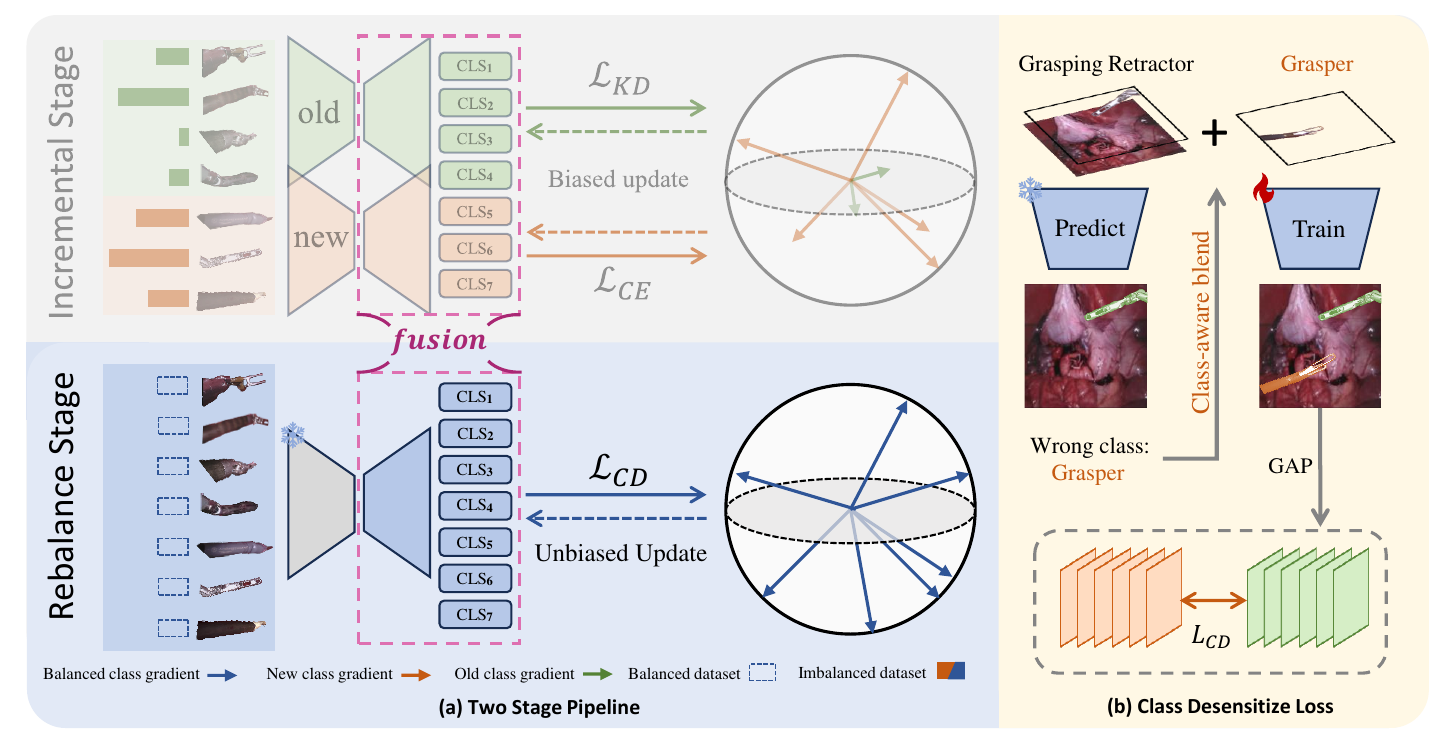}
\caption{Framework of Our Two-Stage Class-Incremental Continual Segmentation Method. Following the incremental training stage, similar to previous approaches, we introduce (a) an additional rebalance stage using a synthetic balanced dataset. During this stage, (b) a Class Desensitization Loss is applied to reduce class bias introduced in the incremental stage. A Dynamic Weight Fusion, shown in the purple dotted box, is performed between the biased model and the unbiased model to integrate knowledge derived from more diverse real-world data.
 } 
\label{fig:pipeline}
\end{figure*}

The proposed two-stage method addresses a continual learning scenario relevant to real-world clinical applications, where models must adapt incrementally to new classes after deployment. The training process is divided into two stages, as illustrated in Figure~\ref{fig:pipeline}.  In the incremental stage, similar to existing methods, the process begins with a model that has already been trained on prior classes at step \( t-1 \). The goal is to incrementally update the model to recognize a new class at step \( t \) while preserving its performance on previously learned classes. We introduce an additional rebalance stage to address the class imbalance problem caused during incremental training. In this stage, the model from the incremental stage undergoes a post-training process specifically designed to further enhance performance. Unlike conventional approaches, the rebalance stage balances weight updates while keeping the backbone parameters frozen. This is achieved using a synthetic dataset for fine-tuning, ensuring that core feature extraction remains unaltered. By introducing the rebalance stage, our method effectively addresses challenges in class-incremental learning, significantly improving robustness and stability in robotic surgical settings.

\subsection{Problem Formulation: Incremental Stage}
Let us define the dataset at each step as $\mathcal{D}_{t} = \{x_i, y_i\}$, where $x_i$ represents the data and $y_i$ is the associated label. The label space for the training task $T$ is given as $C_t \cup \{c_b\}$, where $C_t$ contains all classes involved in the task, and $c_b$ corresponds to the background. Typically, classification loss is handled using cross-entropy loss. Due to the disjoint nature of class sets across tasks, i.e., $C_i \cap C_j = \varnothing$, the objectives differ significantly between tasks, often causing catastrophic forgetting. To mitigate this, knowledge distillation is a widely adopted method. In continual learning, distillation is generally categorized into two types: feature-based distillation and logits-based distillation, which can be expressed as:
\begin{align}
\mathcal{L}_{FD} &= \frac{1}{|\mathcal{D}_{t}|} \sum_{(x_i, y_i) \sim \mathcal{D}_{t}} \left\| \Psi_{old}(x_i) - \Psi_{new}(x_i) \right\|^2 \\
\mathcal{L}_{LD} &= \frac{1}{|\mathcal{D}_{t}|} \sum_{(x_i, y_i) \sim \mathcal{D}_{t}} KL \left( \Phi_{old}(\Psi_{old}(x_i)), \Phi_{new}(\Psi_{new}(x_i)) \right)
\end{align}
\(\Psi_{old}\)/\(\Phi_{old}\) and \(\Psi_{new}\)/\(\Phi_{new}\) denote the feature extractor/classifier from the previous old model and incremental new model, respectively, and \(D\) is the corresponding dataset. 

The softmax cross-entropy loss is used as the classification loss, which is computed as follows: 
\begin{equation}
\mathcal{L}_{CE} = \frac{1}{|\mathcal{D}_{t}|}\sum_{(x_i, y_i) \sim \mathcal{D}_t} -y_i \log \left[ \Phi_{new}(x_i) \right]    
\end{equation}
$\Phi_{new}(x_i)$ is the output probability of new classifier. 

This approach introduces a significant bias toward new classes due to the imbalance between the abundant samples of new classes and the limited exemplars from old classes. During training, the current model starts with the weights of the previous model, with an additional output channel added to handle segmentation for the new class. These class-specific parameters are highly prone to overfitting on the current class. By examining the weights of the segmentation head for both new and old classes, as illustrated in Figure~\ref{fig:biased_head}, it is evident that the bias in class predictions results from uneven weight updates. Our work aims to address and correct the bias in the segmentation head.

\subsection{Proposed Method: Rebalance Stage}
Building on our finding that the segmentation head is heavily biased, we propose a simple and effective bias correction method. Our approach introduces additional training stages, as illustrated in Figure~\ref{fig:pipeline}. The basic idea is to alleviate the bias by learning a balanced distribution of both old and new classes (shown in Fig~\ref{fig:pipeline}). First, we train the feature extractor and segmentation head using the baseline method. In the rebalance stage, we freeze the feature extractor and fine-tune the segmentation head on a synthetic, class-balanced set with a class-desensitizing loss, using a few epochs to address the imbalance between new and old classes through balanced updates. After fine-tuning, we apply a DWF between the original and fine-tuned weights to integrate prior knowledge, enhancing the model's generalization ability. This section details the generation of the synthetic class-balanced set, the class-desensitizing loss, and the DWF method.

\subsubsection{Inpaint and Blending}
A balanced dataset is essential for rebalancing and removing bias. However, in incremental surgical segmentation, acquiring such a dataset is challenging. We pioneer the integration of inpainting and blending techniques into the continual learning setting. 
Specifically, we utilize an inpainting model to generate background images. 
The foreground instrument is cropped separately and then overlaid onto the generated background images to address data limitations.
We carefully control the sample class distribution to avoid class imbalance, effectively addressing issues of data scarcity and reducing the need for costly, time-intensive labeling.

In surgical settings, backgrounds are highly valuable, but due to privacy concerns and operational constraints, obtaining and storing a large number of background images is not feasible. Available samples are often coupled with specific foreground categories, resulting in an unbalanced dataset. Although recent efforts have used generative models for data synthesis, these models often struggle with high-quality generation due to the detailed nature of surgical backgrounds and the similarity of many surgical instruments in the foreground~\citep{xu2024privacy}. By using inpainting, we can reuse existing backgrounds and flexibly create a balanced dataset. Compared to direct generative models, inpainting provides better detail preservation as it retains more of the original information. 

\begin{figure}[!t]
\centering
\includegraphics[width=\linewidth]{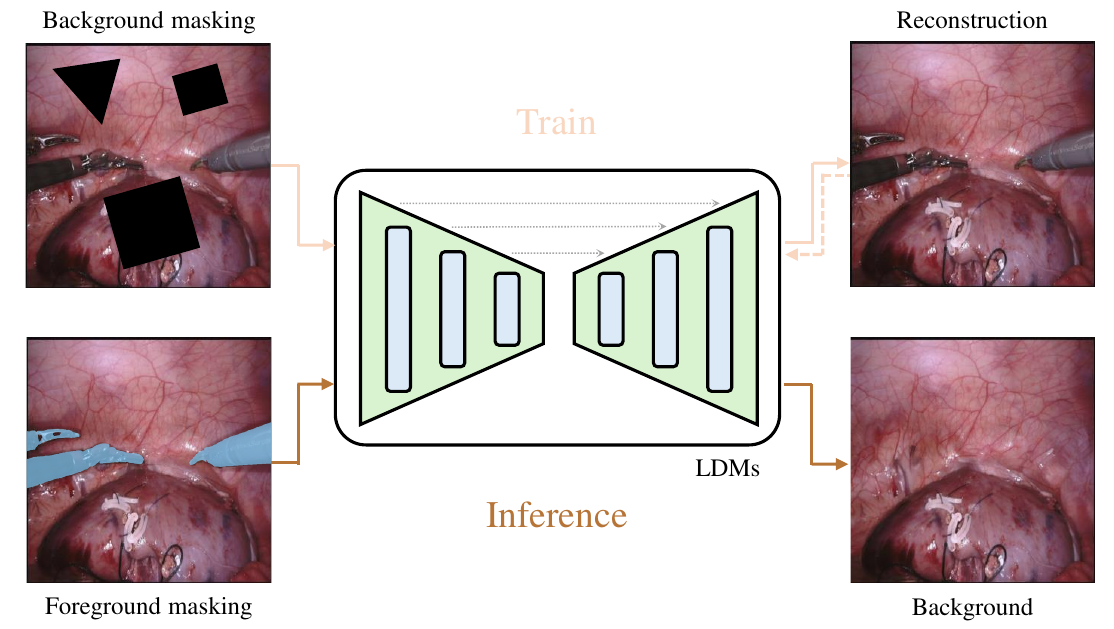}
\caption{Construction of a rebalance dataset through inpainting: During training, the model performs a reconstruction task on background-masked regions. At inference time, it is used to remove existing foreground instruments to isolate the background. } 
\label{fig:inpaint}
\end{figure}

Given the remarkable performance demonstrated by diffusion models recently, we chose to use latent diffusion models (LDM) to generate the background images~\citep{rombach2022high}. However, robotic surgical images differ significantly from natural images, so directly applying existing models is not ideal due to domain gaps. To address this, we fine-tune the pre-trained LDM on our surgical dataset. Specifically, we first fine-tune the pre-trained LDM autoencoder, and then train the U-Net component. During training, we mask out the surgical background regions in the images, allowing the model to learn the background inpainting process. Subsequently, during inference, the inpainting model is used to remove the foreground surgical instruments from the given data samples, generating clean background images. The inpainting data samples are illustrated in Figure~\ref{fig:inpaint}. The dataset is denoted as $\mathcal{D}_r = \{\mathcal{B}, \mathcal{F} \}$, where $\mathcal{B}$ represents the background obtained through inpainting, and $\mathcal{F} = \{f^{i}\}$ denotes the foreground instrument, with each $f^{i}$ corresponding to different class $i$.

In the surgical setting, various environmental disturbances can lead to performance deviations between training and test datasets. Insufficient model robustness under these conditions may result in prediction errors, potentially compromising procedural safety~\citep{garcia2021image,wang2023generalizing}. To address this, we enhance both backgrounds and foregrounds through augmentations such as rotation, scaling, and flipping. In this procedure, we allow up to a specified number of surgical instruments to appear simultaneously in a single background image. We then blend the augmented foreground and background images to generate composite images along with corresponding multi-class masks.

\subsubsection{Class Desensitization Loss}

After obtaining clean background images, our next challenge is to use these backgrounds for model training in a way that allows for balanced updates, reducing bias, and improving performance. Additionally, it is also a challenge to avoid introducing a gap between synthetic and real data. As demonstrated in FIGURE~\ref{fig:biased_head}, the average summation and t-SNE have shown that the segmentation heads have a strong bias toward certain classes.

Motivated by contrastive learning~\citep{michieli2021continual, wu2023continual}, we design the Class Desensitization Loss (CDL) as our objective function, employing contrastive learning with class-aware blending to mitigate the bias introduced in incremental stages. Contrastive learning supports representation learning for both old and new classes by enhancing intra-class compactness and inter-class separation. By enabling the model to learn more robust and discriminative feature representations, CDL also addresses biases that might arise from the synthetic-real domain shift by acting as a corrective mechanism.

We use our existing backgrounds and instrument foregrounds to perform class-aware blending. First, we randomly select an instrument foreground and blend it into the background according to the blending method described above. Then, we use the existing model to predict the label of this foreground. As previously discussed, biased models tend to misclassify certain classes as other similar classes, which may lead to errors in category prediction when applied to the initial blended image. Based on the predicted label, we select a foreground of the predicted class and blend it again. The resulting samples are used for training and loss computation, as shown in Fig~\ref{fig:pipeline}(b). The overall algorithm of CDL is shown in Algorithm~\ref{algorithm}.

\renewcommand{\algorithmicrequire}{\textbf{Input:}}
\renewcommand{\algorithmicensure}{\textbf{Output:}}

\begin{algorithm}[ht]
    \small 
    \caption{Class Desensitization Loss}
    \label{algorithm}
    \begin{algorithmic}[1]
        \REQUIRE Model $\mathcal{F}$; rebalanced dataset $\mathcal{D}_{\text{rebalance}} = \{\mathcal{B}, \mathcal{F} \}$; blending number $N$
        \ENSURE Class Desensitization Loss $\mathcal{L}_{\text{CD}}$

        \STATE Randomly select a foreground object $f^i_{0}$ and a background $b_0$ from dataset $\mathcal{D}$
        \STATE Blend $f^i_{0}$ with $b_0$ to create a blended image $I_0$ and assign it label $Y_0$
        \STATE Use model $\mathcal{F}$ to predict the label $\hat{Y}_0$ for the foreground in $I_0$
        \IF {$\hat{Y}_0$ is the background class}
            \STATE Set the new blending class $C \gets i$
        \ELSIF {$\hat{Y}_0 \neq Y_0$}
            \STATE Set the new blending class $C \gets c$, where $c$ is the incorrectly predicted class
        \ENDIF
        \FOR {$j = 1$ to $N$}
            \STATE Randomly select a foreground object $f^C_{j}$
            \STATE Blend $f^C_j$ into $I_{j-1}$ to create a blended image $I_j$ with label $Y_j$
        \ENDFOR
        \STATE Compute the loss $\mathcal{L}_{\text{CD}}(\mathcal{F}(\mathrm{I}_j), Y_j)$
    \end{algorithmic}
\end{algorithm}

The samples contain different classes that can confuse the model, which can be addressed by the CDL using contrastive learning, defined as:
\begin{align}
\mathcal{L}_{CD} &= \frac{1}{2} \left( y \cdot \|\mathbf{x}_1 - \mathbf{x}_2\|_2^2 + (1 - y) \cdot \max(0, m - \|\mathbf{x}_1 - \mathbf{x}_2\|_2^2) \right) \\
\mathcal{L}_{CE} &= - \sum_{c=1}^{C} Y \log(\hat{Y}) \\
\mathcal{L}_{BCE} &= - [Y \log(\hat{Y}) + (1 - Y) \log(1 - \hat{Y})]
\end{align}
where \(\mathbf{x}_1\) and \(\mathbf{x}_2\) represent the features of the confused classes, computed via global averaging. The variable \(y\) serves as the label to indicate the positive or negative class, \(Y\) and \(\hat{Y}\) are the predicted probabilities and ground truth, and \(m\) represents the margin for the contrastive difference. This loss operates alongside the conventional pixel-wise cross-entropy loss, \(L_{CE}\), and the binary cross-entropy loss, \(L_{BCE}\), offering a complementary approach to improve segmentation performance. The combination is intentional: while pixel-wise cross-entropy loss focuses on predicting the correct label for each sample, CDL targets effective data representation by considering the relationships between different samples.

Thus, the training objective is to optimize the following combined loss:
\begin{equation}
    \mathcal{L} = \alpha \mathcal{L}_{CD} + \beta (\mathcal{L}_{CE} + \mathcal{L}_{BCE})
\end{equation}
where \(\alpha\), \(\beta\) represent the respective weights for each loss component.

\begin{figure}[!t]
\centering
\includegraphics[width=\linewidth]{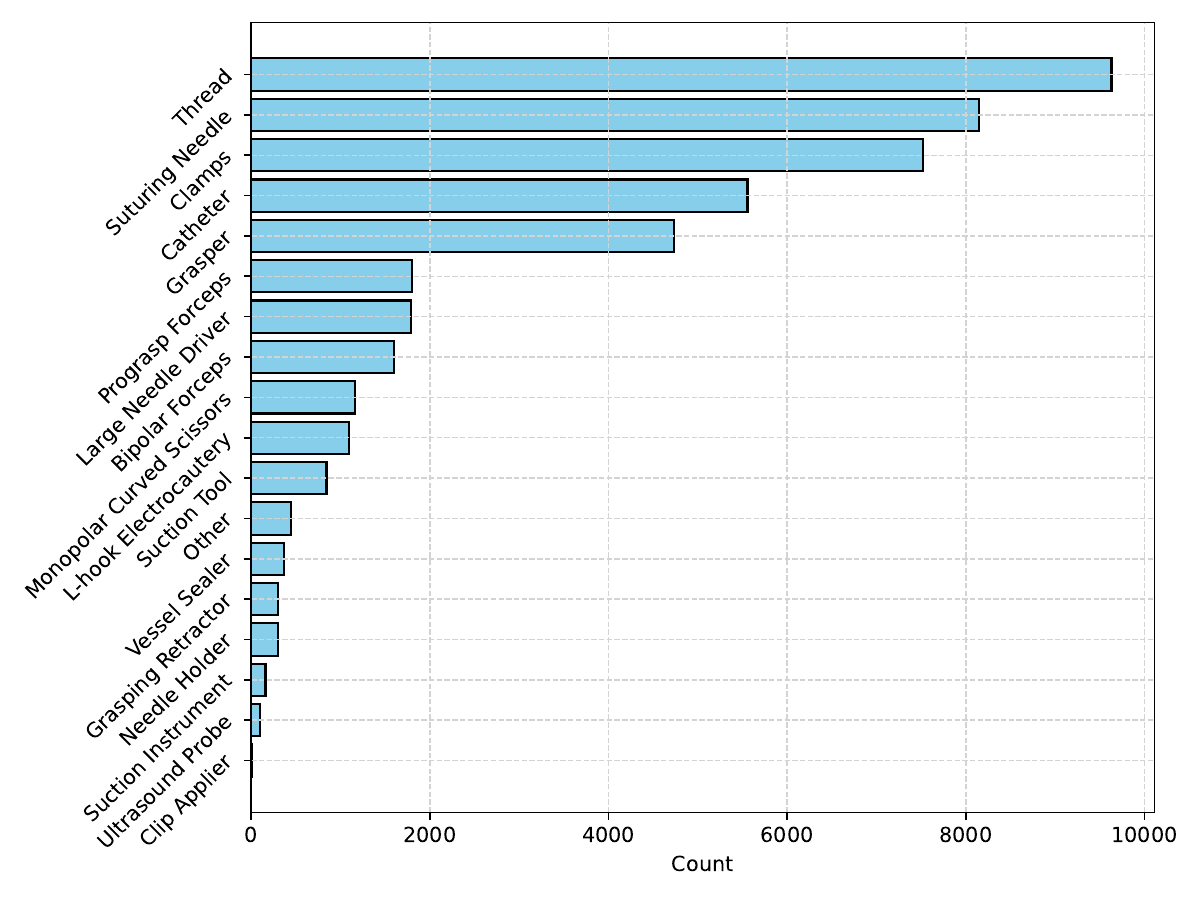}
\caption{Illustration of the class distribution in the collected dataset, ordered from the most frequent to the least frequent class.} 
\label{fig:dataset_distribution}
\end{figure}

\subsubsection{Dynamic Weight Fusion}
The primary concept of our Dynamic Weight Fusion (DWF) is to enhance the model's generalization capability while maintaining balanced representations. Although the incremental model may exhibit a significant bias toward certain classes, it often encodes more detailed knowledge about these classes. This is because the training dataset during the incremental stage is typically more diverse than the one used in the rebalance stage. To effectively retain general knowledge while preserving an unbiased representation, we propose the DWF strategy to further improve performance. Specifically, after the rebalance stage, we have two models: one trained during the incremental stage (\(\theta_{\text{ori}}\)) and another after the rebalancing process (\(\theta_{\text{re}}\)).  We introduce an additional parameter \(\lambda\), which serves as a weight fusion factor, enabling us to combine the parameters of these two models in a specified ratio. This ratio acts as a balancing factor. The operation for dynamic weight fusion can thus be expressed as:
\begin{equation}
\theta_{\text{fused}} = \lambda \cdot \theta_{\text{ori}} + (1 - \lambda) \cdot \theta_{\text{re}}
\end{equation}
This formulation ensures that the model benefits from both the detailed knowledge retained in \(\theta_{\text{ori}}\) and the balanced representation provided by \(\theta_{\text{re}}\), allowing for improved performance while mitigating biases. While \(\lambda\) serves as a control ratio for parameter magnitudes, we also introduce a constraint on the selection of fusion parameters to mitigate the bias present in \(\theta_{\text{ori}}\). To address this, we define \(\alpha\), a dynamic threshold parameter, based on the most variable parameters after rebalancing. These parameters are likely to represent general knowledge learned from a large dataset, which is at risk of being lost during rebalancing. The threshold \(\alpha\) is computed as follows:
\begin{equation}
\alpha = \min(\text{top-}k(|\theta_{\text{ori}} - \theta_{\text{re}}|)),
\end{equation}
where \(\text{top-}k(|\theta_{\text{ori}} - \theta_{\text{re}}|)\) identifies the \(k\)-largest differences between \(\theta_{\text{ori}}\) and \(\theta_{\text{re}}\), capturing the parameters most affected by rebalancing.
The overall strategy can be expressed as:
\begin{equation}
\theta_{\text{fused}} = 
\begin{cases} 
\theta_{\text{ori}}, & \text{if } |\theta_{\text{ori}} - \theta_{\text{re}}| < \alpha, \\
(1 - \lambda)\theta_{\text{ori}} + \lambda\theta_{\text{re}}, & \text{if } |\theta_{\text{ori}} - \theta_{\text{re}}| \geq \alpha,
\end{cases}
\end{equation}
indicating that weight fusion is applied selectively to parameters likely to be significantly influenced by class imbalance. This targeted fusion ensures the preservation of general knowledge while addressing the issues introduced during the incremental training stage.

\section{Experiments}
\subsection{Implementation}
\subsubsection{Datasets}
We collected a new dataset to assess our method. The motivation behind collecting the new dataset stems from the limitations identified in previous datasets, which include a restricted number of instrument classes and the challenge of class imbalance in real-world surgical workflows. In surgical environments, some instruments are used more frequently than others, leading to highly imbalanced data distributions. Addressing this imbalance is crucial for evaluating the robustness of Continual Learning (CL) models under realistic conditions, where the model must adapt to new classes without access to past data (a scenario prone to catastrophic forgetting).

\begin{figure}[t!]
\centering
\includegraphics[width=\linewidth]{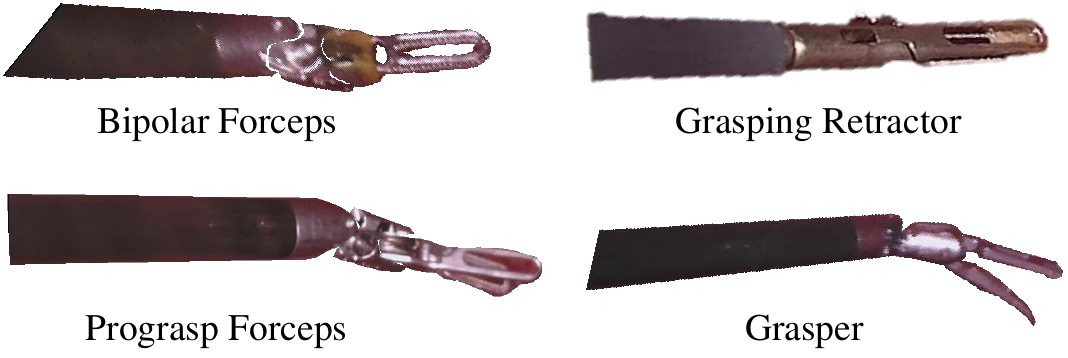}
\caption{Demonstration of visually similar samples across different classes in our dataset. We present several surgical instruments with similar appearances, including Bipolar Forceps, Prograsp Forceps, Grasping Retractor, and Grasper.}
\label{fig:dataset_similarity}
\end{figure}

Our dataset is merged from four public datasets: \textit{EndoVis 2017 Dataset}~\citep{allan20192017}, \textit{EndoVis 2018 Dataset}~\citep{allan20202018}, \textit{CholecSeg8k}~\citep{hong2020cholecseg8k}, and \textit{SAR-RAPR50}~\citep{psychogyios2023sar}. The \textit{EndoVis 2017 and 2018} datasets primarily focus on the segmentation of robotic instruments in minimally invasive surgeries, offering detailed pixel-wise annotations for da Vinci robotic tools, including articulating parts such as shafts, jaws, and wrists, as well as surgical devices like ultrasound probes and suturing instruments. \textit{CholecSeg8k} provides segmentation data for key instruments used in laparoscopic cholecystectomy, including graspers and electrocautery tools, alongside complex anatomical backgrounds. \textit{SAR-RARP50} contains segmentation annotations for robotic instruments in the context of radical prostatectomy surgeries. Our newly collected dataset expands the number of instrument classes compared to previous datasets, as shown in Figure~\ref{fig:dataset_distribution}. The integrated dataset contains 18 instrument classes, with a highly imbalanced distribution. It includes 66 videos with a total of 26,638 frames, of which 19,933 are in the training set and 6,705 are in the test set. The resolution ranges from 1920 to 800 pixels, originating from RGB endoscopic videos. Each frame was annotated instance by instance, with separate annotated masks provided for each object.
Additionally, the dataset includes several visually similar classes, resembling real robotic surgical scenes. FIGURE~\ref{fig:dataset_similarity} presents examples of categories with similar appearances.

\renewcommand{\arraystretch}{1.1}  
\begin{table*}[t]
\centering
\caption{Performance comparison across different continual learning tasks (13-1, 13-5, 9-3, and 17-1, ``x-y" denotes that x is the number of old classes and y is the number of incremental classes introduced in each step). Results are reported for methods without continual learning (WO CL), including the old model at time point \( t = 0 \) and naive fine-tuning (FT), and those with continual learning (W CL). The metrics presented are for base, new, and all classes, with the offline training serving as an upper bound.}
\resizebox{\textwidth}{!}{
\begin{tabular}{l|ccc|ccc|ccc|ccc}
\hline 
\multirow{2}{*}{\textbf{Method}} & \multicolumn{3}{c|}{\textbf{13-1 (5 steps)}} & \multicolumn{3}{c|}{\textbf{13-5 (1 step)}} & \multicolumn{3}{c|}{\textbf{9-3 (3 steps)}} & \multicolumn{3}{c}{\textbf{17-1 (1 step)}} \\ 
\cline{2-13}
    & \textit{Base} & \textit{New} & \textit{All} & \textit{Base} & \textit{New} & \textit{All} & \textit{Base} & \textit{New} & \textit{All} & \textit{Base} & \textit{New} & \textit{All} \\ 
    \hline
Old Model (t=0) &61.29 &0.00 &45.16 &61.29 &0.00 &45.16 &60.62 &0.00 &31.90 &56.93 &0.00 &53.93 \\
FT &0.00 &17.67 &8.53 &0.00 &32.39 &13.71 &0.00 &10.99 &9.52 &0.00 &5.49 &8.77 \\ 
\hline
EWC~\citep{kirkpatrick2017overcoming} &0.00 &10.98 &5.30 &0.00 &\textbf{52.60} &22.25 &0.00 &7.65 &8.75 &0.00 &5.47 &8.73 \\
LwF~\citep{li2017learning} &1.48 &21.82 &11.30 &14.04 &14.23 &20.11 &3.56 &27.42 &18.35 &23.14 &4.64 &23.31 \\
LwF-MC~\citep{rebuffi2017icarl} &0.32 &17.48 &9.05 &10.07 &13.49 &14.35 &5.63 &19.31 &16.02 &10.62 &15.18 &15.98 \\
ILT~\citep{michieli2019incremental} &0.92 &21.70 &11.23 &16.10 &23.03 &23.16 &10.77 &20.04 &19.01 &35.51 &7.13 &35.77 \\
MiB~\citep{cermelli2020modeling} &7.98 &\textbf{25.21} &17.11 &13.56 &19.41 &19.52 &30.79 &7.86 &24.98 &35.04 &7.45 &36.79 \\
RCIL~\citep{zhang2022representation} &50.08 &4.86 &40.64 &26.22 &48.53 &42.66 &0.38 &22.95 &22.64 &52.32 &8.41 &50.59 \\
CAT-SD~\citep{xu2024privacy} &44.70 &6.29 &37.32 &49.50 &11.24 &41.45 &4.95 &5.55 &9.67 &55.31 &9.00 &52.58 \\
MBS~\citep{park2024mitigating} & 30.85& 7.64 & 24.40&46.93&40.54&45.15&35.92&18.16&34.12&46.43&49.41&46.59 \\
NeST~\citep{xie2024early} & 29.61& 24.08 & 28.08&39.59&30.05&37.89&18.56&8.26&13.41&45.64&9.44&43.63 \\
\cline{1-13}
PLOP~\citep{douillard2021plop} &12.09 &19.69 &18.49 &44.14 &23.38 &41.47 &4.56 &18.37 &14.90 &39.11 &8.32 &41.06 \\
\rowcolor[HTML]{EAF6FC}
\textbf{PLOP+Ours} &52.50 &18.74 &45.99 &55.02 &23.37 &\textbf{48.93} &34.63 &11.94 &28.33 &57.19 &\textbf{19.25} &54.43 \\
\rowcolor[HTML]{EAF6FC}
\textit{Improvement} & \textit{\textcolor[HTML]{D75555}{+40.41}} & \textit{\textcolor[HTML]{93D557}{-0.95}} & \textit{\textcolor[HTML]{D75555}{+27.50}} & \textit{\textcolor[HTML]{D75555}{+10.88}} & \textit{\textcolor[HTML]{93D557}{-0.01}} & \textit{\textcolor[HTML]{D75555}{+7.46}} & \textit{\textcolor[HTML]{D75555}{+30.07}} & \textit{\textcolor[HTML]{93D557}{-6.43}} & \textit{\textcolor[HTML]{D75555}{+13.43}} & \textit{\textcolor[HTML]{D75555}{+18.08}} & \textit{\textcolor[HTML]{D75555}{+10.93}} & \textit{\textcolor[HTML]{D75555}{+13.37}} \\
\cline{1-13}
EWF~\citep{xiao2023endpoints} &54.82 &5.12 &43.98 &50.90 &26.83 &43.22 &35.96 &0.00 &23.77 &57.15 &0.05 &53.93 \\
\rowcolor[HTML]{EAF6FC}
\textbf{EWF+Ours} &\textbf{62.37} &8.70 &\textbf{50.13} &\textbf{60.83} &20.18 &45.62 &\textbf{63.96} &18.80 &\textbf{46.68} &\textbf{59.01} &8.88 &\textbf{55.33} \\
\rowcolor[HTML]{EAF6FC}
\textit{Improvement} & \textit{\textcolor[HTML]{D75555}{+7.55}} & \textit{\textcolor[HTML]{D75555}{+3.58}} & \textit{\textcolor[HTML]{D75555}{+6.15}} & \textit{\textcolor[HTML]{D75555}{+9.93}} & \textit{\textcolor[HTML]{93D557}{-6.65}} & \textit{\textcolor[HTML]{D75555}{+2.40}} & \textit{\textcolor[HTML]{D75555}{+28.00}} & \textit{\textcolor[HTML]{D75555}{+18.80}} & \textit{\textcolor[HTML]{D75555}{+22.91}} & \textit{\textcolor[HTML]{D75555}{+1.86}} & \textit{\textcolor[HTML]{D75555}{+8.83}} & \textit{\textcolor[HTML]{D75555}{+1.40}} \\

\hline
Offline &58.00 &60.43 &60.77 &58.00 &60.43 &60.77 &58.00 &60.43 &60.77 &58.00 &60.43 &60.77 \\
\hline 
\end{tabular}}
\label{table:all_res}
\end{table*}

\renewcommand{\arraystretch}{1.1}  
\setlength{\tabcolsep}{4.5pt}  
\begin{table*}[t]
\centering
\caption{Performance comparison across different datasets for the 13-1 continual learning task. Results are reported for methods without continual learning (WO CL), including the old model at time point \( t = 0 \) and naive fine-tuning (FT), and those with continual learning (W CL). The metrics include performance on base, new, and all classes, with offline training serving as the upper bound.}
\resizebox{\textwidth}{!}{
\begin{tabular}{l|ccc|ccc|ccc|ccc}
\hline 
\multirow{2}{*}{\textbf{Method}} & \multicolumn{3}{c|}{\textbf{CholecSeg8k}} & \multicolumn{3}{c|}{\textbf{EndoVis 2017}} & \multicolumn{3}{c|}{\textbf{EndoVis 2018}} & \multicolumn{3}{c}{\textbf{SAR-RAPR50}} \\ 
\cline{2-13}
    & \textit{Base} & \textit{New} & \textit{All} & \textit{Base} & \textit{New} & \textit{All} & \textit{Base} & \textit{New} & \textit{All} & \textit{Base} & \textit{New} & \textit{All} \\ 
    \hline
Old Model (t=0) & 62.53 & 0.00 & 44.66 & 83.82 & 0.00 & 55.88 & 73.33 & 0.00 & 61.10 & 61.69 & 0.00 & 52.97 \\
FT & 0.00 & 0.00 & 31.60 & 0.00 & 0.00 & 11.00 & 0.00 & 0.00 & 12.02 & 0.00 & 75.05 & 24.49 \\ 
   \cline{1-13}
EWC~\citep{kirkpatrick2017overcoming} & 0.00 & 0.00 & 41.26 & 0.00 & 0.00 & 16.63 & 0.21 & 0.97 & 12.91 & 0.00 & \textbf{74.82} & 23.45 \\
LwF~\citep{li2017learning} & 0.00 & \textbf{78.39} & 58.30 & 0.00 & 0.44 & 11.78 & 0.00 & 0.05 & 12.81 & 0.00 & 66.23 & 22.69 \\
LwF-MC~\citep{rebuffi2017icarl} & 0.00 & 18.40 & 37.86 & 0.00 & 0.00 & 11.03 & 0.00 & 0.00 & 12.14 & 0.00 & 19.56 & 16.55 \\
ILT~\citep{michieli2019incremental} & 0.00 & 59.66 & 51.94 & 0.00 & 0.00 & 11.26 & 0.00 & 0.00 & 12.24 & 0.00 & 36.75 & 18.70 \\
MiB~\citep{cermelli2020modeling} & 47.41 & 15.80 & 31.60 & 22.00 & 2.75 & 11.00 & 14.49 & 6.21 & 12.42 & 16.15 & 20.93 & 20.93 \\
RCIL~\citep{zhang2022representation} & 66.63 & 31.19 & 65.31 & 44.07 & 21.47 & 41.91 & 38.50 & 22.34 & 43.85 & 51.03 & 0.00 & 49.75 \\
CAT-SD~\citep{xu2024privacy} & 42.22 & 53.63 & 64.44 & 33.74 & 3.26 & 30.36 & 22.55 & 2.13 & 30.35 & 37.06 & 33.73 & 43.82 \\
MBS~\citep{park2024mitigating}& 27.20&34.53&41.52&35.54&10.83&37.12&18.93&10.16&23.12&27.96&18.32&23.21 \\
NeST~\citep{xie2024early}& 31.23&29.53&40.95&28.79&13.37&29.62&18.02&12.61&25.49&28.28&23.77&26.42 \\

\cline{1-13}
PLOP~\citep{douillard2021plop} & 0.00 & 68.52 & 55.65 & 14.59 & 6.18 & 21.80 & 13.16 & 6.90 & 24.15 & 0.00 & 43.28 & 18.61 \\ 
\rowcolor[HTML]{EAF6FC}
\textbf{PLOP+Ours} & 59.70 & 45.69 & 67.86 & 46.28 & 67.98 & 60.84 & \textbf{63.68} & 25.55 & 62.91 & 11.55 & 12.66 & 23.82 \\ 
\rowcolor[HTML]{EAF6FC}
\textit{Improvement} & \textit{\textcolor[HTML]{D75555}{+59.70}} & \textit{\textcolor[HTML]{93D557}{-22.83}} & \textit{\textcolor[HTML]{D75555}{+12.21}} & \textit{\textcolor[HTML]{D75555}{+31.69}} & \textit{\textcolor[HTML]{D75555}{+61.80}} & \textit{\textcolor[HTML]{D75555}{+39.04}} & \textit{\textcolor[HTML]{D75555}{+50.52}} & \textit{\textcolor[HTML]{D75555}{+18.65}} & \textit{\textcolor[HTML]{D75555}{+38.76}} & \textit{\textcolor[HTML]{D75555}{+11.55}} & \textit{\textcolor[HTML]{93D557}{-30.62}} & \textit{\textcolor[HTML]{D75555}{+5.21}} \\

\cline{1-13}
EWF~\citep{xiao2023endpoints} & 63.46 & 67.46 & 76.46 & \textbf{57.65} & 31.25 & 52.73 & 51.67 & 41.99 & 56.82 & 49.53 & 0.01 & 48.99 \\
\rowcolor[HTML]{EAF6FC}
\textbf{EWF+Ours} & \textbf{67.12} & 70.04 & \textbf{78.57} & 50.31 & 48.05 & \textbf{55.36} & 58.90 & \textbf{49.07} & \textbf{63.00} & \textbf{51.68} & 0.02 & \textbf{50.84} \\ 
\rowcolor[HTML]{EAF6FC}
\textit{Improvement} & \textit{\textcolor[HTML]{D75555}{+3.66}} & \textit{\textcolor[HTML]{D75555}{+2.58}} & \textit{\textcolor[HTML]{D75555}{+2.11}} & \textit{\textcolor[HTML]{93D557}{-7.34}} & \textit{\textcolor[HTML]{D75555}{+16.80}} & \textit{\textcolor[HTML]{D75555}{+2.63}} & \textit{\textcolor[HTML]{D75555}{+7.23}} & \textit{\textcolor[HTML]{D75555}{+7.08}} & \textit{\textcolor[HTML]{D75555}{+6.18}} & \textit{\textcolor[HTML]{D75555}{+2.15}} & \textit{\textcolor[HTML]{D75555}{+0.01}} & \textit{\textcolor[HTML]{D75555}{+1.85}} \\

   \hline
Offline & 64.68 & 74.97 & 79.59 & 64.47 & 65.33 & 59.10 & 50.11 & 56.55 & 49.56 & 77.91 & 67.15 & 58.76 \\
\hline 
\end{tabular}}
\label{table:dataset_res}
\end{table*}

\subsubsection{Experiments Setting}
In our surgical instrument segmentation scenario, we assume that the training data from the time point(t=0) is no longer accessible during the subsequent incremental learning stages. We define four tasks: 13-1 (5 steps), 13-5 (1 step), 9-3 (3 steps), and 17-1 (1 step), all set under the overlapped protocol. ``x-y" denotes that x is the number of old classes and y is the number of incremental classes introduced in each step. At each step, the model has access to the new dataset, but previous datasets are not available for further training. Similar to the overlapped protocol described earlier, the training set at each stage may include images containing pixels belonging to classes that are yet to be learned in future steps, with these pixels labeled as background.  This setting allows us to evaluate the effectiveness of the incremental learning approach when handling new instrument classes while dealing with the limitations imposed by inaccessible prior data.

\subsubsection{Implementation Details}
We used the DeepLabv3+ architecture~\citep{chen2018encoder} with the ResNet-101 backbone~\citep{he2016deep} for segmentation tasks. We compared our proposed method against the following baselines and state-of-the-art solutions: EWC~\citep{kirkpatrick2017overcoming}, LwF~\citep{li2017learning}, LwF-MC~\citep{rebuffi2017icarl}, ILT~\citep{michieli2019incremental}, MiB~\citep{cermelli2020modeling}, RCIL~\citep{zhang2022representation}, CAT-SD~\citep{xu2024privacy}, MBS~\citep{park2024mitigating}, NeST~\citep{xie2024early}, PLOP~\citep{douillard2021plop}, and EWF~\citep{xiao2023endpoints}.
Consistent with prior approaches, in-place activated batch normalization was set in the backbone. All input images were resized to {224$\times$224} for uniformity. Using the SGD optimizer, we set the initial learn rate as 0.06 and 0.006 for the continual learning steps, and the batch size is set to 12. For the incremental stage, we set the training to 20 epochs per step. Taking the 13-1 (5 steps) setting as an example, PLOP requires approximately 6~h for training, and other baseline methods also require similar training time. The rebalance stage is set to 1 step with 3 epochs, which takes around 0.5~h. The training of the rebalance stage requires around 9~GB of VRAM. All experiments are conducted on two NVIDIA RTX 3090 GPUs.

The model testing and weight initialization followed the same approach as outlined earlier. Specifically, the best-performing model from each step was selected based on the highest mean Intersection over Union (IoU) achieved on the validation set. For each continual learning step, the model was initialized with the weights of the previously trained model, ensuring that knowledge from earlier steps was retained. During the continual learning step, the old model acted as a teacher, providing logits for distillation loss calculation, while the model was optimized using both cross-entropy loss and distillation losses. This approach enabled us to learn new classes incrementally while maintaining performance in previously learned classes.

\begin{figure*}[!t]
\centering
\includegraphics[width=\linewidth]{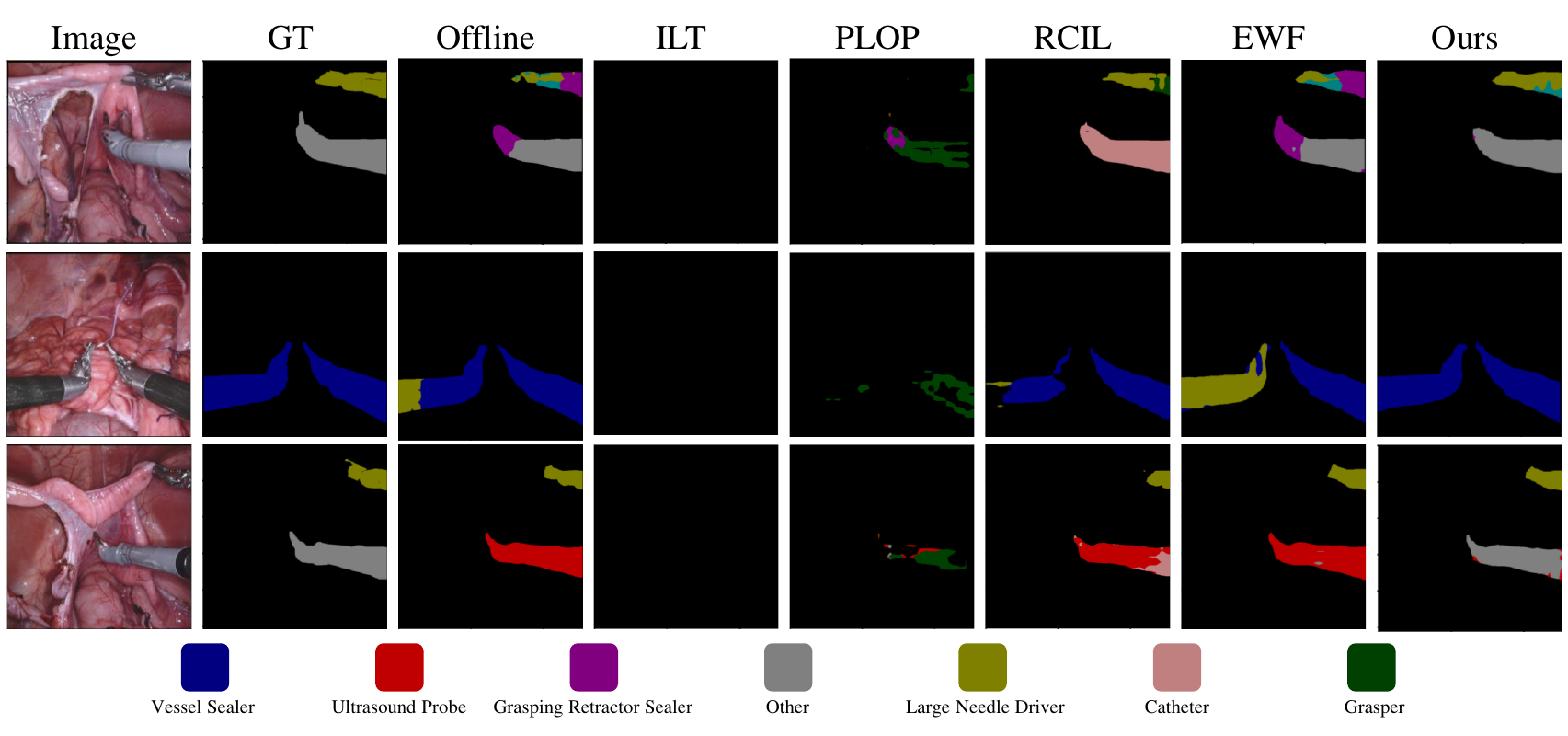}
\caption{Visual comparison with different state-of-the-art methods (Task 13-1 on our surgical dataset).} 
\label{fig:vis}
\end{figure*}

\subsection{Experimental Results}
We conduct extensive experiments and evaluations to validate our proposed approach and investigate the effectiveness of privacy-preserving continual learning in robotic surgery. Our method is compared against several state-of-the-art continual learning approaches, which can be categorized into four groups. One category includes approaches based on penalty computation, such as EWC~\citep{kirkpatrick2017overcoming}, which employ strategies to compute the importance of parameters for old classes, safeguarding those critical parameters to prevent forgetting. The second category consists of methods based on knowledge distillation, originally designed for continual image classification, including LwF~\citep{li2017learning} and LwF-MC~\citep{rebuffi2017icarl}. These methods have demonstrated impressive performance when applied to semantic segmentation tasks. The third category focuses on approaches specifically designed for continual semantic segmentation, such as ILT~\citep{michieli2019incremental}, which combines both feature- and output-level knowledge distillation, and MiB~\citep{cermelli2020modeling}, which relies solely on output-level knowledge distillation to address the background shift problem. Recently, some weight fusion methods have been proposed to improve knowledge transfer and enhance model robustness, including RCIL~\citep{zhang2022representation} and EWF~\citep{xiao2023endpoints}. Additionally, we record the results of the fine-tuning approach (FT) as the lower bound for segmentation performance, since FT updates the model with new classes without any additional constraints, leading to the most severe catastrophic forgetting. The upper bound is represented by the results from the offline training approach (Offline), where images and annotations for all classes are available during training. To ensure a fair comparison, we implement all competitors using the same base segmentation network, as well as identical computing environments and common hyperparameter settings.

In TABLE~\ref{table:all_res} and TABLE~\ref{table:dataset_res}, we present a statistical comparison of our method with existing approaches across tasks 13-1, 13-5, 9-3, and 17-1 on our surgical dataset, as well as task 13-5 on the four-component dataset. Our observations indicate that, regardless of the method employed, old class forgetting remains a challenge when learning new classes. Furthermore, learning new classes can be negatively impacted when methods impose excessive constraints. While weight fusion methods maintain base class performance, they limit the plasticity required for learning new classes. By addressing the class bias between base and new classes, our method not only preserves the knowledge of base classes but also enhances their performance. For instance, on the 13-1 setting, our algorithm achieves performance gains of 27.5\% and 6.15\% mIoU for PLOP and EWF, respectively, by effectively mitigating bias. This mitigation further improves the performance of the base class by 40.41\% and 7.55\%, respectively, while the slight decrease in the performance of the new class may be attributed to the reduction in predictable pixels. The differences in improvements between the two baseline methods are primarily because PLOP~\citep{douillard2021plop} relies on feature distillation to limit representation capacity, while EWF~\citep{xiao2023endpoints} employs a more robust weight fusion technique. The distinction in their principles leads to differences in plasticity, which results in varying degrees of performance improvement. Notably, as continual learning progresses, the model updates increasingly favor new classes, resulting in greater bias. This trend further highlights the effectiveness of our method in addressing such challenges. FIGURE~\ref{fig:biased_head} provides evidence for the bias mitigation capabilities of our proposed framework. The increased uniformity in the distribution of the summation average and t-SNE embeddings demonstrates a significant reduction in class bias.

These findings underscore the robustness and effectiveness of our approach in mitigating bias, preserving base class knowledge, and improving overall performance across diverse tasks.
Fig~\ref{fig:vis} shows the visual comparison with the offline and partial competitors. We observe that our method consistently provides better predictions for Vessel Sealer, Other, Ultrasound Probe, Grasping Retractor Sealer, Large Needle Driver, Catheter, and Grasper, each represented by different colors. Additionally, our predictions are comparable to those of the best-performing offline method, as evidenced by fewer instances of over-predicting and under-predicting, as well as reduced confusion among different instrument classes.

\begin{table}[t]
\centering
\caption{Ablation study on the Class Desensitization Loss and dynamic weight fusion in our Rebalance Stage.}
\label{table:abl}
\renewcommand{\arraystretch}{1.1}  
\setlength{\tabcolsep}{4.5pt}  
\resizebox{0.48\textwidth}{!}{
\begin{tabular}{c|cc|c c c c}
\hline
Setting & $\mathcal{L}_{CD}$ & \makecell[c]{WF} & 13-1  & 13-5 & 9-3 & 17-1 \\
\hline
Incremental Stage& -   &-   &43.98   &43.22  &23.77  &53.93 \\
\hline
\multirow{3}{*}{\textbf{Rebalance Stage}}
& \ding{55}   &\ding{55}   &45.88   &45.29 &41.38 &50.09 \\
& \ding{51}   &\ding{55}   &49.83   &43.06 &45.75 &54.64 \\
\rowcolor[HTML]{EAF6FC}
\cellcolor[HTML]{FFFFFF}& \ding{51}   &\ding{51}   &\textbf{50.13}   &\textbf{45.62}  &\textbf{46.68}  &\textbf{55.33} \\

\hline
\end{tabular}}
\end{table}

\begin{figure*}[!t]
\centering
\hspace{-0.3cm}
\includegraphics[width=0.6\linewidth]{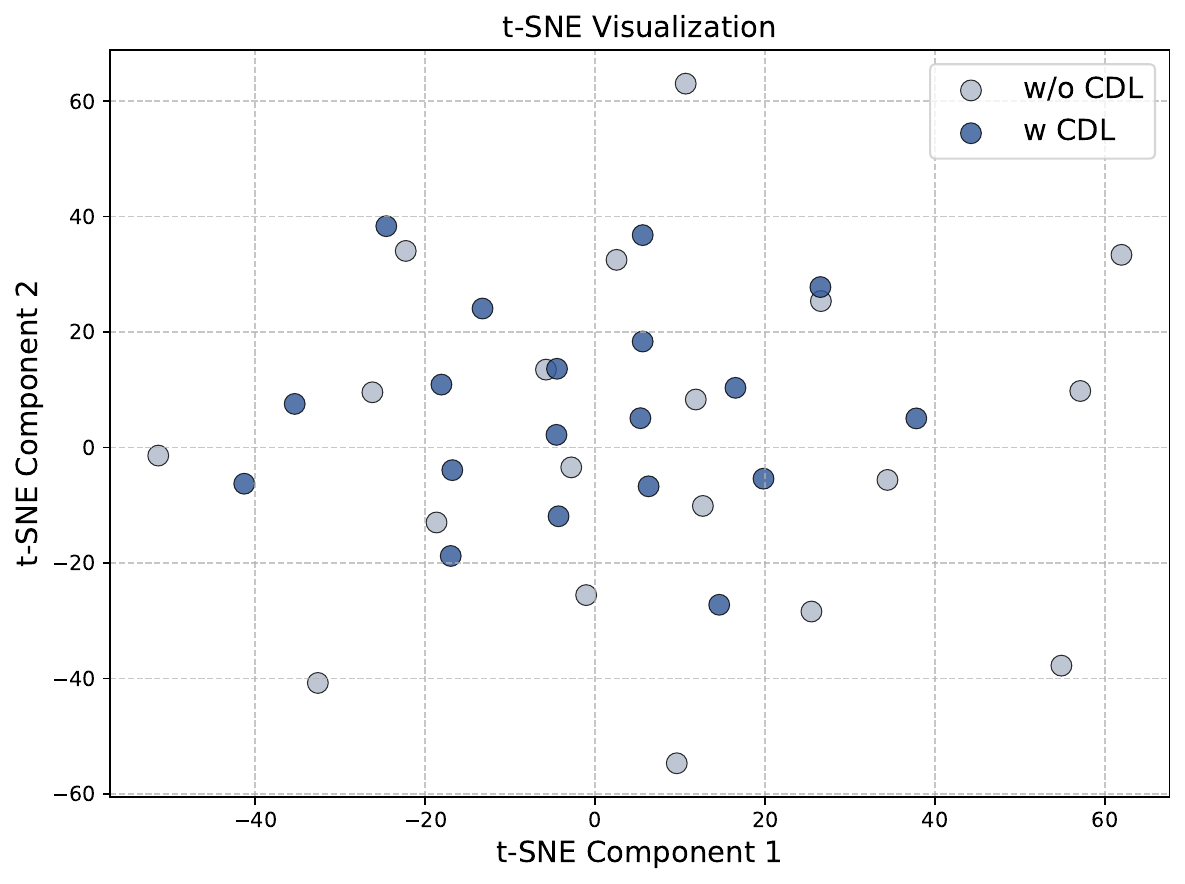}
    \caption{Ablation study on the Class Desensitization Loss via the t-SNE visualization on the feature space.}
\label{fig:t-SNE_CDL}
\end{figure*}

\subsection{Ablation Study}
In this part, we demonstrate and analyze the effectiveness of the Class Desensitization Loss and its blending method. We also provide an analysis of our Dynamic Weight Fusion Strategy. We use 13-1 (5 steps) for the ablation experiments. 
TABLE~\ref{table:abl} presents the ablation study on the impact of each proposed module, including the rebalance stage, Class Desensitization Loss, and Dynamic Weight Fusion. With the inclusion of the rebalance stage, performance improves by 1.90\%, 2.07\%, and 17.61\% in the 13-1, 13-5, and 9-3 tasks, respectively. This demonstrates that a simple bias adjustment step can significantly enhance the performance of continual segmentation models. However, in the 17-1 task, a slight decrease in performance is observed, primarily due to the fewer continual learning steps, where incremental classes have not introduced significant class bias. This effect is mitigated by the Class Desensitization Loss, which results in a 4.55\% improvement in the 17-1 task by clarifying the class boundary. We also perform a quality analysis to ensure that the bias has been mitigated, as demonstrated by the t-SNE feature in FIGURE~\ref{fig:t-SNE_CDL}, accompanied by qualitative cluster metrics in TABLE~\ref{table:t-SNE_CDL}. The DB Index, CH Index, and WCSS have significantly improved, showing a more uniform distribution, which serves as evidence of the efficiency. Additionally, the Dynamic Weight Fusion strategy proves to be a versatile method for combining previous models, enhancing the generalization ability on real samples. Ultimately, our method strikes a balance between enabling the learning of new classes and maintaining the ability to distinguish base classes.

\begin{table}[t]
\centering
\caption{Ablation study on the Class Desensitization Loss using cluster metrics on the t-SNE embedding.}
\label{table:t-SNE_CDL}
\renewcommand{\arraystretch}{1.1}  
\setlength{\tabcolsep}{4.5pt}  
\begin{tabular}{c|c c c c}
\hline 
$\mathcal{L}_{CD}$  & DB Index $\downarrow$ & CH Index $\uparrow$ & WCSS $\downarrow$ \\
\hline \ding{55}   &3.56   &1.13   &31445.56  \\
\hline
\rowcolor[HTML]{EAF6FC}\ding{51}& \textbf{2.13} &  \textbf{3.02}  &\textbf{11366.84} \\

\hline
\end{tabular}
\end{table}

\begin{table}[t]
\centering
\caption{Comparisons of our ablation study on different blending methods and sample numbers.}
\label{table:tw0-stage_number}
\renewcommand{\arraystretch}{1.1}  
\setlength{\tabcolsep}{4.5pt}  
\resizebox{0.48\textwidth}{!}{
\begin{tabular}{c|c|c|c|c}
\hline
Method      & Number & Base classes & New classes & All classes \\
\hline
\multirow{4}{*}{Random} 
            & 100    & 44.20        & 2.64        & 36.02       \\
            & 300    & 48.39        & 5.43        & 39.66       \\
            & 500    & 50.83        & 7.69        & 41.93       \\
            & 1000   & 54.08       &7.33        & 44.05      \\
\hline
\multirow{4}{*}{\textbf{Class-aware}} 
            & 100    & 46.65        & 7.02        & 38.78       \\
            & 300    & 49.38        & 4.50        & 40.09       \\
            & 500    & 58.77        & 6.21        & 46.97       \\
            \rowcolor[HTML]{EAF6FC}
\cellcolor[HTML]{FFFFFF}
            & 1000   & \textbf{62.02}        & \textbf{8.49}        & \textbf{49.83}       \\
\hline
\end{tabular}}
\end{table}

\begin{table}[t]
\centering
\caption{Comparisons of different background synthetic methods. Crop refers to directly cropping the background from samples.}
\label{table:background}
\renewcommand{\arraystretch}{1.1}  
\setlength{\tabcolsep}{4.5pt}  
\resizebox{0.48\textwidth}{!}{
\begin{tabular}{c|c|c|c}
\hline
Method & Base classes & New classes & All classes \\
\hline
Crop & 56.10 & 6.47 & 45.21 \\
Generate (DDPM,~\citealp{ho2020denoising}) & 44.02 & 5.79 & 36.49 \\
\rowcolor[HTML]{EAF6FC}
\cellcolor[HTML]{FFFFFF}\textbf{Inpaint} (LDM,~\citealp{rombach2022high}) & \textbf{62.37} & \textbf{8.70} & \textbf{50.13}\\
\hline
\end{tabular}}
\end{table}

\begin{table}[t]
\centering
\caption{Synthesis quality comparison of different generation methods.}
\label{table:syn_methods}
\renewcommand{\arraystretch}{1.1}  
\setlength{\tabcolsep}{4.5pt}  
\resizebox{0.45\textwidth}{!}{
\begin{tabular}{c|ccc}
\hline
Method & FID $\downarrow$ & IS $\uparrow$ & PSNR $\uparrow$ \\
\hline
DDPM~\citep{ho2020denoising} & 266.77 &0.99 & 11.25 \\
StyleGAN-XL~\citep{sauer2022stylegan} & 138.46 & 1.09 & 12.16\\
FLUX.1-Fill-dev~\citep{flux2024} & 158.60 & \textbf{1.86} & 11.06\\

\rowcolor[HTML]{EAF6FC}
LDM~(\textbf{Ours},~\citealp{rombach2022high}) & \textbf{48.26} & 1.30 & \textbf{12.40} \\
\hline
\end{tabular}}
\end{table}

\begin{figure*}[!t]
\centering
\hspace{-0.3cm}
\includegraphics[width=\linewidth]{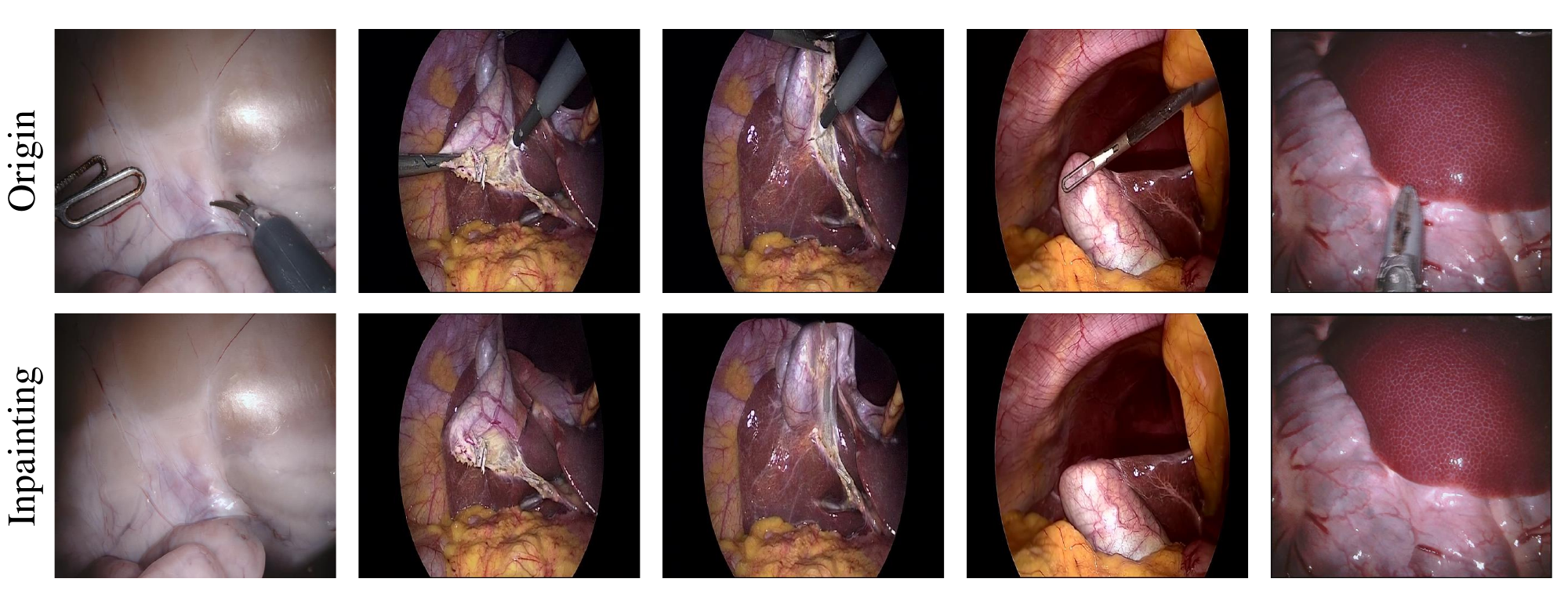}
\caption{Visualization samples of our synthetic background inpainting.}
\label{fig:inpaint_vis}
\end{figure*}
\subsubsection{Class-aware Blending}
In TABLE~\ref{table:tw0-stage_number}, we demonstrate the performance of different blending methods and sample numbers. We compare random blending with class-aware blending, verifying the effectiveness of our algorithm in alleviating model bias caused by class imbalance through the use of synthetic class samples. The results show that smaller sample sizes (e.g., 100 samples) can negatively impact model performance, particularly for new classes. However, as the number of samples increases, the performance improves significantly, with the class-aware method outperforming random blending. The class-aware blending method leads to a better balance between base and new classes, resulting in enhanced overall performance.

\begin{figure*}[!t]
\centering
\includegraphics[width=0.8\linewidth]{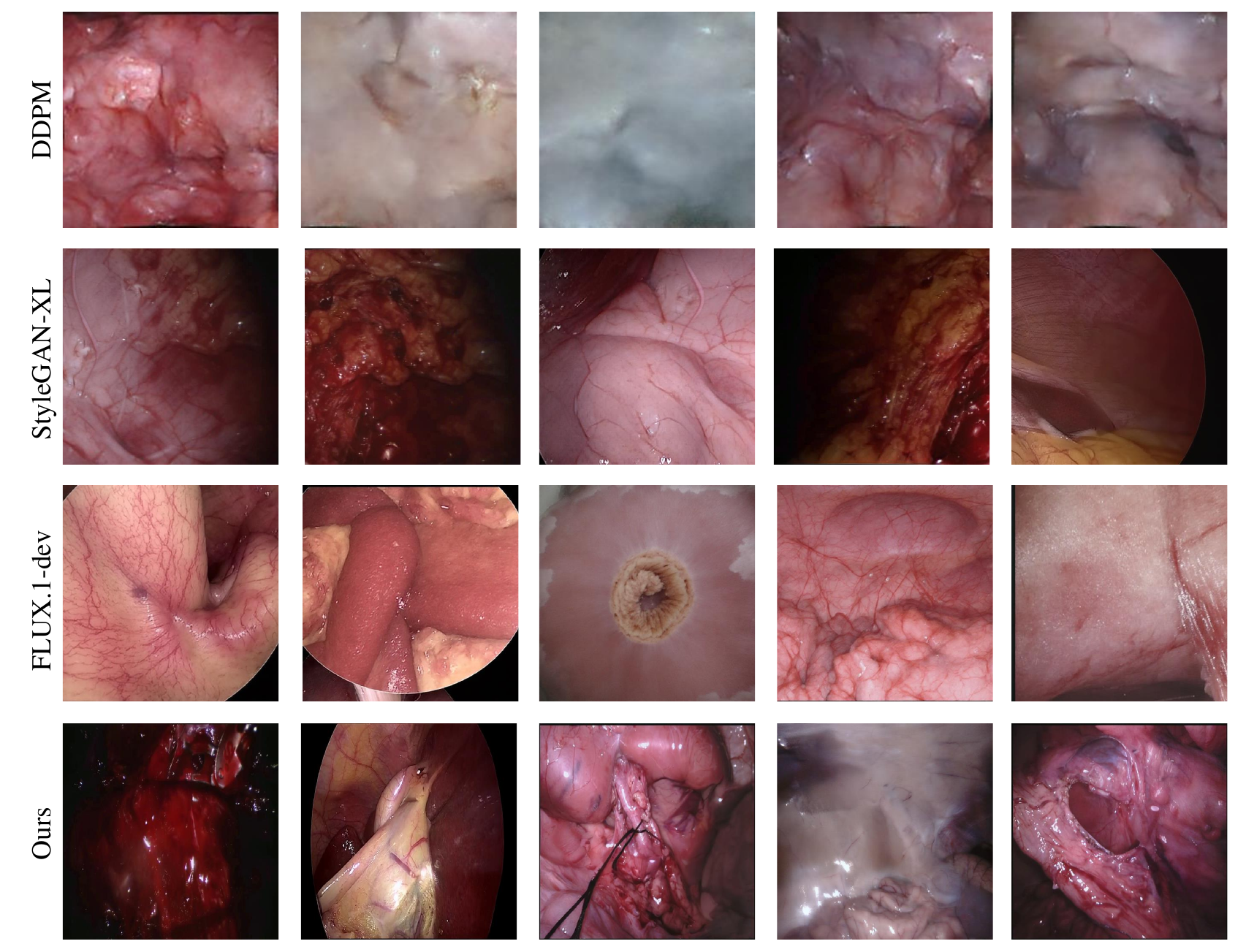}
\caption{Qualitative comparison of different generation methods.} 
\label{fig:contrast_syn}
\end{figure*}

TABLE~\ref{table:background} illustrates that directly cropping the background from the sample would reduce the effectiveness of the Class Desensitization Loss enhancement. 
Moreover, in TABLE~\ref{table:syn_methods}, we explore the synthesis quality of four generative models (DDPM,~\citealp{ho2020denoising}, StyleGAN-XL,~\citealp{sauer2022stylegan}, FLUX.1-Fill-dev,~\citealp{flux2024}, and LDM,~\citealp{rombach2022high}) applied to our inpainting task and finally select LDM~\citep{rombach2022high} as our inpainting model.
LDM exhibits strong performance compared to other methods, as other methods mostly rely on cropped patches instead of using the latent space, thus losing completeness. 
FIGURE~\ref{fig:inpaint_vis} presents visualization examples of our inpainting solution, and FIGURE~\ref{fig:contrast_syn} provides the qualitative comparison of different synthetic methods. Additionally, the performance of the generation method significantly degrades due to its low quality. In contrast, our inpainting method demonstrates a substantial performance improvement. Besides, TABLE~\ref{table:loss_weight} shows the performance comparison under different loss weight configurations, specifically varying the values of $\alpha$ and $\beta$. When $\alpha = 1.0$ and $\beta = 1.0$, the model performs best on base classes and overall classes, leading to an improved performance of 3.96\%.

In summary, our class-aware blending strategy combines foreground and background through a simple overlay. While this approach may introduce certain data inconsistencies and biases due to the lack of effective modeling between foreground and background, it has already proven effective in improving the performance of downstream segmentation tasks. Our work does not focus on developing generative models specifically for surgical applications. Instead, the generative model can serve as a simple and effective solution to address the downstream continual segmentation task in the surgical domain. Furthermore, our generation-based approach can also enhance privacy protection and improve data scalability~\citep{wang2022rethinking,wang2023generalizing}.

\subsubsection{Fusion Strategy}
In TABLE~\ref{table:weight_fusion}, we demonstrate the performance of different fusion strategies and their corresponding parameter selections. The Replace method involves directly selecting the top-\(k\) parameter values from the previous model to rebalance the current model. However, this approach results in a performance decline, as it can reintroduce biased parameters into the model. For example, when top-\(k=1\%\), the performance drops to 45.88\% for all classes, with a notable decrease in base class accuracy (55.98\%). As \(k\) increases (e.g., top-\(k=50\%\)), the performance continues to degrade further (43.80\% overall), illustrating the detrimental effect of bringing biased parameters back into the model. In contrast, our Weighted Fusion strategy balances the knowledge from the previous model while maintaining the model’s state unbiased. This method shows a clear improvement over the Replace strategy. Specifically, when using top-\(k=10\%\) and \(\lambda=0.01\), the model achieves the highest overall performance at 50.13\%, with a 0.3\% improvement over the baseline (49.83\%). The weighted fusion approach also allows for better retention of both base and new class performance, as evidenced by the slight improvements in both categories. Notably, limiting the number of fused parameters (e.g., top-\(k=10\%\)) helps prevent performance degradation, keeping the model less prone to bias. These results emphasize the effectiveness of the weighted fusion strategy in enhancing model performance, offering a robust solution for continual learning scenarios where knowledge from previous models must be incorporated without introducing bias.

\begin{table}[t]
\centering
\caption{Comparisons of our ablation study on the weight of the Class Desensitization Loss (Task 13-1 on our surgical dataset). }
\label{table:loss_weight}
\renewcommand{\arraystretch}{1.1}  
\setlength{\tabcolsep}{4.5pt}  
\begin{tabular}{cc|c|c|c}
\hline
$\alpha$ & $\beta$ & Base classes & New classes & All classes \\
\hline
1.0 & 0.0 & 56.54 & 7.76 & 45.92 \\
0.0 & 1.0 & 61.33 & \textbf{9.90} & 49.65 \\
\rowcolor[HTML]{EAF6FC}
\cellcolor[HTML]{FFFFFF}\textbf{1.0} & \cellcolor[HTML]{FFFFFF}\textbf{1.0} & \textbf{61.98} & 8.74 & \textbf{49.84} \\
1.0 & 2.0 & 59.98 & 5.79 & 47.74 \\
2.0 & 1.0 & 61.65 & 7.82 & 49.36 \\
2.0 & 2.0 & 57.22 & 4.64 & 45.48 \\
\hline
\end{tabular}
\end{table}

\begin{table}[h]
\centering
\caption{Comparisons of our ablation study on the fusion parameter selection top-\(k\) and merged weight \(\lambda\) (Task 13-1 on our surgical dataset). }
\label{table:weight_fusion}
\renewcommand{\arraystretch}{1.1}  
\setlength{\tabcolsep}{4.5pt}  
\resizebox{0.48\textwidth}{!}{
\begin{tabular}{c|cc|c|c|c}
\hline
Method & top-$k$ &$\lambda$  & Base classes & New classes & All classes \\
\hline
None & - & - & 62.02 & 8.49 & 49.83 \\
\hline
\multirow{4}{*}{Replace} 
& 1\% & - & 55.98 & 9.31 & 45.88 \\
& 10\% & - & 56.38 & 5.87 & 45.25 \\
& 30\% & - & 54.76 & 5.32 & 44.00 \\
& 50\% & - & 54.69 & 5.05 & 43.80 \\
\hline
\multirow{5}{*}{\textbf{Weighted}} 
& 10\% & 0.1 & \textbf{62.41} & 8.13 & 49.97 \\
& 10\% & 0.3 & 62.34 & 8.11 & 49.92 \\
& 30\% & 0.05 & 61.34 & 7.94 & 49.19 \\
& 50\% & 0.1 & 58.56 & 7.16 & 47.09 \\
\rowcolor[HTML]{EAF6FC}
\cellcolor[HTML]{FFFFFF}
& 10\% & 0.01 & 62.37 & \textbf{8.70} & \textbf{50.13} \\
\hline
\end{tabular}}
\end{table}

\section{Discussion and Conclusion}
In this work, we propose a novel two-stage pipeline to address the critical challenge of class imbalance in continual segmentation for robotic surgical scenes, a challenge exacerbated by privacy constraints and the complexity of surgical workflows. 
Our SurgCSS framework includes the class-aware blending with inpainting to generate a class-balanced dataset by separating instrument foregrounds from surgical backgrounds, alongside the CDL based on contrastive learning to reduce confusion between similar classes. Additionally, a DWF strategy combines incremental and rebalance models to enhance class representation and improve overall segmentation accuracy.
Experimental results demonstrate that (i) there is significant performance degradation due to class imbalance when directly applying conventional segmentation techniques; (ii) existing continual learning methods, while alleviating catastrophic forgetting, fail to address class imbalance effectively in surgical contexts; and (iii) our two-stage pipeline successfully balances performance across old and new classes, outperforming state-of-the-art continual learning methods with average 11.82\% improvement in robotic surgical scenarios. 

This is a fundamental work focusing on resolving the combined issues of class imbalance, catastrophic forgetting, and privacy concerns in the context of robotic surgical segmentation. The framework can be extended to other medical imaging tasks (e.g., organ segmentation or lesion detection), as these tasks often share similar challenges such as imbalanced class distributions and insufficient data. To address these challenges, our proposed inpainting synthetic data generation method and Class Desensitization Loss offer flexible, generalized solutions that reduce class bias and improve model robustness across various applications in the medical domain.
However, our proposed framework still has some limitations:
(i) Although the final deployed segmentation model remains identical to the previous model without introducing additional parameters, it still requires offline data generation and extra training for the generative model, which significantly increases the training cost. (ii) Any defects or artifacts in the synthetic data can negatively affect the model's performance, particularly when distinguishing between visually similar instrument classes. (iii) The generative model, trained on the background characteristics of the initial dataset, may require retraining or substantial updates to effectively generate data for significantly different surgical domains, leading to extra computational costs. Therefore, future work should rigorously validate the feasibility, robustness, and applicability of the proposed continual learning approach and generative framework in real-world clinical environments, including addressing lightweight deployment and real-time inference challenges in actual hospital settings.
Additionally, our continual learning approach can be effectively applied to adapt visual foundation models, such as SAM~\citep{kirillov2023segment}, to various downstream medical applications, mitigating catastrophic forgetting of the originally learned computer vision patterns.

\vspace{-3mm}
\section*{Acknowledgment}

This work was supported by Hong Kong Research Grants Council GRF 14204524, GRF 14203323, GRF 14216022, GRF 14211420, CRF C4026-21GF; Guangdong Basic \& Applied Research Fund Regional Joint Fund Project 2021B1515120035 (B.02.21.00101); Shenzhen-Hong Kong-Macau Technology Research Programme (Type C) STIC Grant SGDX20210823103535014 (202108233000303).

\section*{Declaration of Using Generative AI and AI-assisted Technologies}
The authors have used ChatGPT to improve the readability and language of the work. The process of using ChatGPT was done with human oversight and control. The authors have carefully reviewed and edited the results. The authors have not used ChatGPT to provide citations.

\bibliographystyle{elsarticle-harv} 
\bibliography{mybib}{}

\end{document}